\documentclass[final,5p,times,twocolumn]{elsarticle}

\usepackage{placeins}
\usepackage{rotating}
\usepackage{longtable} % for 'longtable' environment
\usepackage{pdflscape}
\usepackage{xcolor}
\usepackage{color}
\usepackage{multirow}
\usepackage[caption=false]{subfig}
\usepackage{array}
\usepackage{hyperref} 

\newcolumntype{C}[1]{>{\centering\let\newline\\\arraybackslash\hspace{0pt}}m{#1}}

\usepackage{amssymb}
 \usepackage{amsthm}
 \usepackage{amsmath}
 \usepackage{multicol}
\usepackage{graphicx}
\usepackage{booktabs}
\usepackage{multirow}
\usepackage{siunitx}
\usepackage{makecell}
\usepackage{stfloats}
\usepackage{floatrow}
\usepackage{algorithm}% <=================http://ctan.org/pkg/algorithms
\usepackage{algpseudocode}% <========== http://ctan.org/pkg/algorithmicx
\usepackage{hyperref}

\usepackage{algpseudocode}

\newcommand\Algphase[1]{%
\vspace*{-.7\baselineskip}\Statex\hspace*{\dimexpr-\algorithmicindent-2pt\relax}\rule{\textwidth}{0.4pt}%
\Statex\hspace*{-\algorithmicindent}\textbf{#1}%
\vspace*{-.7\baselineskip}\Statex\hspace*{\dimexpr-\algorithmicindent-2pt\relax}\rule{\textwidth}{0.4pt}%
}

\journal{}

\begin{document}

\begin{frontmatter}



\title{Multimodal Ensemble with Conditional Feature Fusion for Dysgraphia Diagnosis in Children from Handwriting Samples}


\author[inst1]{Jayakanth Kunhoth }

\affiliation[inst1]{organization={Department of Computer Science and Engineering , Qatar University},%Department and Organization
            addressline={}, 
            city={Doha},
            postcode={}, 
            state={Qatar},
            country={}}

\author[inst1]{Somaya Al Maadeed}
\author[inst1]{Moutaz Saleh}

\author[inst1]{Younes Akbari}
\begin{abstract}
Developmental dysgraphia is a neurological disorder that hinders children's writing skills. The heterogeneous nature of dysgraphia symptoms and its frequent co-occurrence with other disorders complicate the identification process. In recent years, researchers have increasingly explored machine learning methods to support the diagnosis of dysgraphia based on offline and online handwriting. In most previous studies, the two types of handwriting have been analysed separately, which does not necessarily lead to promising results. In this way, the relationship between online and offline data cannot be explored. To address this limitation, we propose a novel multimodal machine learning approach utilizing both online and offline handwriting data. We created a new dataset by transforming an existing online handwritten dataset, generating corresponding offline handwriting images. We considered only different types of word data (simple word, pseudoword \& difficult word) in our multimodal analysis. We trained SVM and XGBoost classifiers separately on online and offline features as well as implemented multimodal feature fusion and soft-voted ensemble. Furthermore, we proposed a novel ensemble with conditional feature fusion method which intelligently combines predictions from online and offline classifiers, selectively incorporating feature fusion when confidence scores fall below a threshold. Our novel approach achieves an accuracy of 88.8\%, outperforming SVMs for single modalities by 12-14\%, existing methods by 8-9\%, and traditional multimodal approaches (soft-vote ensemble and feature fusion) by 3\% and 5\%, respectively. Our methodology contributes to the development of accurate and efficient dysgraphia diagnosis tools, requiring only a single instance of multimodal word/pseudoword data to determine the handwriting impairment. This work highlights the potential of multimodal learning in enhancing dysgraphia diagnosis, paving the way for accessible and practical diagnostic tools.
\end{abstract}



\end{frontmatter}


%% main text\section{Introduction}\label{sec1}
\section{Introduction}
Recent years have seen an increasing interest in exploring the relationship between neurological disorders, disabilities, and handwriting \cite{drotar2014decision}. This interest has spurred research into using handwriting analysis as a diagnostic tool for various conditions and diseases. Handwriting deterioration, observable in several disorders and disabilities, makes it a promising biomarker. Parkinson's disease, a common neurodegenerative disorder, has been a focal point in handwriting analysis research \cite{drotar2014decision,mucha2018identification,zham2018efficacy,smits2014standardized,nomm2018detailed}, thanks to the availability of several datasets.
Handwriting analysis has demonstrated diagnostic potential for mild cognitive impairment \cite{angelillo2019handwriting}, Alzheimer's disease \cite{stefano2019handwriting}, and disorders within the schizophrenia spectrum and bipolar disorder \cite{crespo2019handwriting}. Dysgraphia, a learning disability affecting writing expression, including spelling, grammar, and the organization of words and letters \cite{Deuel1995}, is closely related to handwriting difficulties. Estimates suggest that 10\% to 30\% of children globally face handwriting challenges. Diagnosing dysgraphia accurately is complex, requiring consideration of various cues that change with age and developmental stage. These signs must persist for at least six months despite intervention efforts \cite{AmericanPsychiatricAssociation.2013}. Dysgraphia can occur alone or with other disorders like autism spectrum disorder (ASD), developmental coordination disorder (DCD), or attention deficit hyperactivity disorder (ADHD), complicating the assessment \cite{Lopez2018}. Early diagnosis and intervention are crucial for effective treatment and reducing effort.

Handwriting analysis data is generally divided into offline and online modes. Researchers examine online handwritten data, recorded in real-time with digitizing tablets, and offline data, comprising scanned or captured handwritten text images \cite{kunhoth2022automated}. Online data features include dynamic parameters such as pen pressure, pen tilt, and pen-tip movement sequences, offering insights into fine motor control and pen dynamics during writing \cite{zham2018efficacy,smits2014standardized,danna2019digitalized,drotar2014analysis,kotsavasiloglou2017machine}. These features help assess detailed character and stroke formation. Advanced features have been introduced to capture more nuanced and condition-specific attributes \cite{drotar2014decision,mucha2018identification,zham2018efficacy,stefano2019handwriting,impedovo2019velocity-based,rios-urrego2019analysis}. Offline data involves processing scanned handwritten text images to extract static features like character shape, size, slant, and spatial distribution on the page. Features related to character and stroke shape and curvature are also extracted to capture handwriting subtleties. Recent studies have focused on extracting CNN features from handwritten images \cite{kunhoth2022automated,KUNHOTH2023120740}.

Despite numerous studies addressing dysgraphia detection through handwriting analysis, most research has separately analyzed online and offline data. Current dysgraphia research in children has mainly emphasized online data analysis, with limited focus on offline data analysis \cite{kunhoth2022automated}. With multimodal analysis at the feature and classifier levels gaining traction and showing improved results in various domains \cite{meng2020survey,rahate2022multimodal,bhunia2020indic}, adopting a multimodal approach in dysgraphia research could enhance diagnosis and treatment outcomes.

To address these significant gaps, our research presents a new dataset created by transforming an existing online handwritten dataset, tailored to evaluate dysgraphia using both offline and online modalities. This dataset aims to offer a more thorough understanding of the condition by capturing its various nuances. Our approach is unique due to its multimodal nature, incorporating both image (offline) data and online handwritten data. This innovative method not only differentiates our research but also expands the potential for dysgraphia diagnosis. Additionally, we have introduced a classifier ensembling method based on conditional feature fusion, which improves performance compared to traditional fusion techniques. Moreover, Our approach can detect whether a child has dysgraphia by analyzing both the online and offline versions of a single written word, unlike most methods in the literature that require analysis of multiple types of handwritten items, including words, sentences, and letters.

The main objectives of this study can be summarized as follows: 

\begin{itemize}
\item Develop and publish a novel multimodal dataset (online and offline handwriting) for automated dysgraphia diagnosis from a single instance of a word written by the subject. Evaluate this dataset using advanced machine learning methods. This transformed dataset is the first publicly available multimodal dataset specifically for dysgraphia diagnosis.
\item Evaluate the performance of each handwriting modality by employing transfer learning for feature extraction on offline data, and extracting kinematic, dynamic, spatial, and temporal features for online data. Develop classifiers using SVM and XGboost algorithms for each modality separately.
\item Assess the significance of multimodal data by implementing both early feature fusion via concatenation of features and late fusion via soft voting-based classifier ensembling. Compare the performance of these fusion methods to highlight the benefits of a multimodal approach.
\item Propose and implement an improved multimodal classifier fusion approach where ensemble of single modality classifiers are conditionally fused with a classifier trained on fused features. It utilizes a prediction confidence threshold to handle uncertain/ near-miss prediction cases in individual modality classifier ensembling approaches, thereby improving diagnostic accuracy.
\end{itemize}

The remainder of the article is organized as follows: Section 2 offers an overview of previous research on the topic. Section 3 introduces the dataset used in this study and details its creation process. Section 4 describes the materials and methods applied, including the specifics of the developed algorithm. In Section 5, we present the evaluation and results. Section 6 provides a comprehensive discussion of the results, limitations, and potential future research directions. Finally, Section 7 concludes the article.

\section{Related works}
The diagnosis of dysgraphia, a learning disability that primarily affects an individual's ability to express themselves through written communication, has been a subject of extensive research in recent years. Despite the inherent complexities involved, numerous studies have explored automated systems leveraging machine learning techniques to aid in the identification and analysis of dysgraphia \cite{kunhoth2022automated}. In the following, we will first look at studies in connection with offline methods and then online methods.

One of the primary approaches has been the analysis of online handwritten data captured through digitizing tablets, aiming to distinguish between typical and dysgraphic handwriting patterns. Mekyska et al. \cite{Mekyska2017} developed methods for classifying handwriting as either typical or dysgraphic by collecting data from school students using a Wacom Intuos tablet. They explored various characteristics such as kinematics, dynamics, and non-linear dynamics to differentiate between typically developing and dysgraphic writing. Their classifiers, based on Random Forest and linear discriminant algorithms, achieved an impressive handwriting classification sensitivity of 96\%.
Richard et al. \cite{Richard2020} evaluated the performance of diverse machine learning algorithms, including Random Forest, logistic regression, and naïve Bayes, in classifying online handwritten features for dysgraphia detection. Their study utilized features like pen tip pressure, letter and word characteristics, including shape and spacing. Asselborn et al. \cite{Asselborn2018} introduced an automated dysgraphia diagnosis tool utilizing a consumer-level tablet and the Ductus software. With 54 extracted handwriting features, including static, kinematic, and dynamic characteristics, their Random Forest (RF) classifier demonstrated excellent accuracy in diagnosing dysgraphia among 298 primary school students, including 56 with dysgraphia.
Drotar et al. \cite{Drotar2020} proposed a machine learning-based system that collected handwriting samples from 120 school students, including those with dysgraphia, using the WACOM Intuos Pro Large tablet. They captured data on pen movement, pressure, azimuth, and altitude during writing. Twenty-two types of spatiotemporal and kinematic features were extracted, and multiple machine learning algorithms were employed for classification, with the AdaBoost algorithm achieving the highest accuracy of 80\%. Notably, features such as pressure and pen lifts showed high discriminatory potential.
A different approach, presented by Dimauro et al. \cite{Dimauro2020}, introduced a software system designed to partially automate the evaluation of the Concise Evaluation Scale for Children's Handwriting (BHK) test. This test involves evaluating thirteen handwriting characteristics and assigning scores based on their quality. The proposed software system automatically generates scores for nine of the thirteen characteristics by modifying multiple document analysis algorithms.
In methods based on online handwriting analysis for dysgraphia diagnosis \cite{Mekyska2017, Asselborn2018, Mekyska2019, Zvoncak2019, dui2020tablet, Gargot2020, Drotar2020, Asselborn2020, KUNHOTH2023104715, 10039584}, spatial characteristics (stroke dimensions and spacing), temporal characteristics (writing time and idle time), dynamic characteristics (pressure, tilt, azimuth), and kinematic characteristics (velocity, acceleration, jerk) play a crucial role in distinguishing normally developing handwriting from dysgraphia.

\begin{figure*}[b]
	\centerline{\includegraphics[width= 14 cm,height=6 cm]{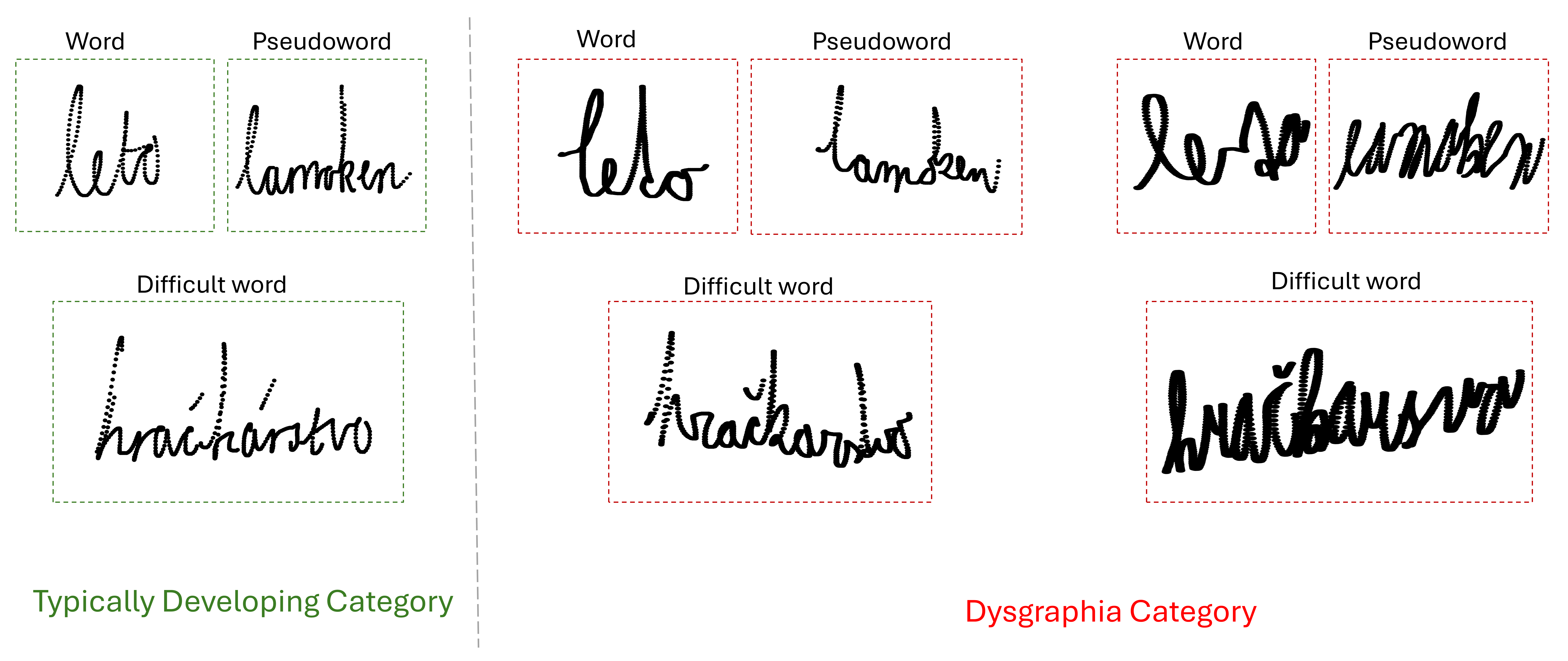}}
	\caption{Handwriting samples of word('leto'), pseudoword('lamoken'), and difficult word('hra\v{c}k'arstvo') from the dataset written by the subjects in each category (typically developing and dysgraphia category. }
	\label{fig_1}
\end{figure*}

In contrast to online data analysis, offline image-based methods focused on extracting various image features from the handwritten product. Devi et al. \cite{devi2022dysgraphia} proposed an end-to-end CNN neural network architecture for classifying images into typically developing and dysgraphic classes. Their research incorporated a combination of handwriting and geometric features, utilizing the Kekre-Discrete Cosine mathematical model to identify dysgraphia. The features were effectively employed in the feature learning stage of deep transfer learning for dysgraphia detection. The Kekre-Discrete Cosine Transform with Deep Transfer Learning (K-DCT-DTL) approach outperformed existing methods, achieving a remarkable accuracy of 99.75\%.
Maurya et al. \cite{10105022} introduced a transfer learning-based approach to discriminate between normal and abnormal handwriting by analyzing images of handwritten letters. They leveraged various pre-trained deep neural network architectures to enhance the accuracy and performance of their handwriting classification system. Kunhoth et al. \cite{KUNHOTH2023120740} proposed a multimodal approach, incorporating transfer learning, deep CNN ensembles, and CNN feature fusion to develop a dysgraphia diagnosis system based on handwritten images. They transformed a publicly available online handwritten dataset into images, covering various writing tasks. Transfer learning using a pre-trained DenseNet201 network produced four task-specific CNN models for words, pseudowords, difficult words, and sentences. These models were combined using soft and hard voting ensembles, while the DenseNet201 network extracted CNN features from each task-specific dataset. The extracted features were fused in different combinations and trained using Support Vector Machines (SVM) classifiers, significantly improving the diagnosis performance.
Despite the extensive research efforts, the literature on machine learning-based dysgraphia diagnosis methods has primarily emphasized online data analysis, with limited exploration of offline handwriting data or images. Notably, the availability of publicly accessible datasets for dysgraphia diagnosis is scarce, with only one identified dataset \cite{Drotar2020} in an online modality. Furthermore, the literature lacks multimodal datasets, indicating a gap in exploring the potential advantages of integrating diverse data sources (online and offline modalities) for a more comprehensive understanding of dysgraphia.

\section{Dataset}

The dataset comprises handwriting samples collected from 57 children diagnosed with dysgraphia and 63 age- and sex-matched controls. The dysgraphic group's average age is 12.25 years (±2.25), while the control group's average age is 11.78 years (±2.09). Handwriting data were captured using a Wacom Intuos Pro Large tablet. Dysgraphic subjects' data were collected during diagnosis by trained professionals, with each subject becoming familiar with the environment and setup beforehand. Control subjects' data were collected in a classroom setting, with similar familiarization opportunities.

The dataset encompasses eight handwriting tasks, including writing the letter 'l' at typical and fast speeds, writing the syllable 'le' at typical and fast paces, and composing words like 'leto' (summer), 'lamoken' (pseudo-word), and the more challenging 'hra\v{c}k'arstvo' (toy-store). The final task involved completing the sentence 'V lete bude teplo a sucho' (The weather in summer is hot and dry) \cite{Drotar2020}. Each sample was assessed by three independent trained professionals to determine its classification into the dysgraphic or healthy control group.

In this research, our approach focused on analyzing various word types ('leto', 'lamoken','hra\v{c}k'arstvo')  excluding letters, syllables, and sentences. This decision aligns with our research objectives, emphasizing the examination of different modalities of word writing data. This online data is converted into images to obtain an offline representation of the data. Each instance was considered as a unique sample for analysis, applied across both online and offline data modalities. However, during evaluation, these multiple instances originating from the same subject are exclusively assigned to either the training or test set, but not distributed across both. This practice ensures that there is no data leakage between the training and test sets, as each subject's instances are used for training or evaluation purposes exclusively, maintaining the integrity of the model assessment. Detailed statistics of samples in the dataset are provided in Table \ref{tab1}.

\begin{table}[]
\caption{Statistics of the multimodal dataset}
\begin{tabular}{|l|ll|}
\hline
\multirow{2}{*}{Handwritten task} & \multicolumn{2}{c|}{Number of samples}                               \\ \cline{2-3} 
                                  & \multicolumn{1}{l|}{Typically developing} & \multicolumn{1}{l|}{Dysgraphia}  \\ \hline
Word                        & \multicolumn{1}{l|}{199}    & \multicolumn{1}{l|}{228}         \\ \hline
Pseudoword                        & \multicolumn{1}{l|}{186}    & \multicolumn{1}{l|}{190}         \\ \hline
Difficult word                    & \multicolumn{1}{l|}{188}    & \multicolumn{1}{l|}{168}         \\ \hline
\end{tabular}
\label{tab1}
\end{table}

\section{Methodology}

\begin{figure*}[b]
	\centerline{\includegraphics[width= 16 cm,height=9 cm]{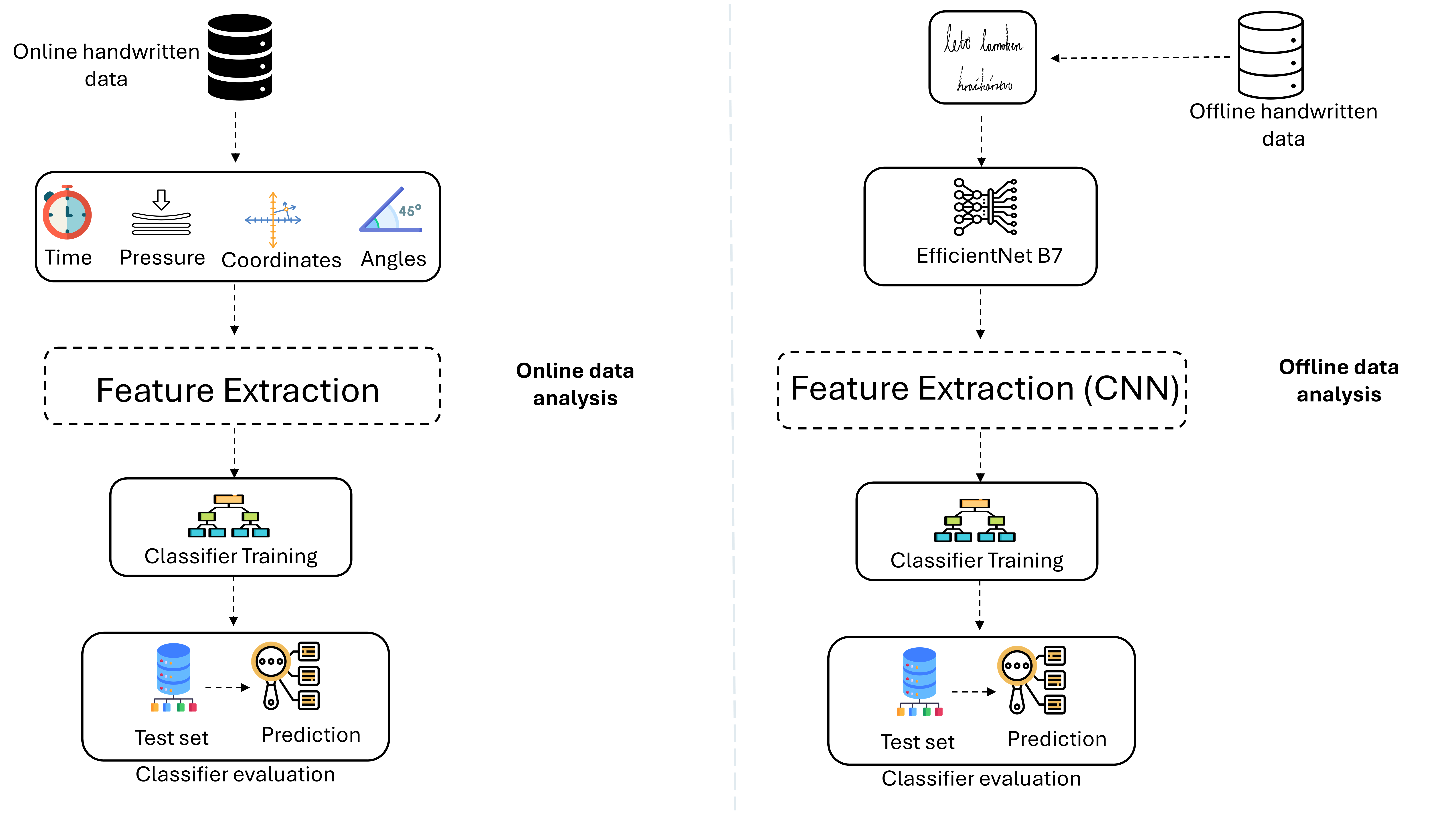}}
	\caption{Dysgraphia diagnosis system development: work flow }
	\label{fig3}
\end{figure*}

This section outlines the methodology employed in constructing classifiers for diagnosing dysgraphia from handwriting samples. Figure \ref{fig3} provides a visual summary of the process for developing a dysgraphia diagnosis system based on handwriting analysis, encompassing both online and offline handwritten data, and detailing the overall workflow.

The development of a machine learning-based dysgraphia diagnosis system involves several key stages. Initially, handwriting experiments are conducted to collect raw data, followed by extracting relevant features that characterize the dynamic, kinematic, temporal, and spatial attributes of the handwritten samples.  With recent technological advancements, the digitized tablet can capture detailed information about the writing process, including the trajectory, time, and dynamics of the pen's movement. 

Moreover, this real-time handwritten data can be transformed into offline data by converting the x and y coordinates into a plot representation. In this study, both online and transformed offline data modalities are considered. Separate feature extraction techniques are applied to each data modality. The extracted features are then compiled into feature vectors that encapsulate the unique characteristics of each subject. These feature vectors are subsequently used to train machine learning algorithms. In this work we utilized SVM and XGboost algorithm to train and evaluate our proposed single modality and multimodality schemes. 

\subsection{Feature extraction}

Feature extraction is crucial in developing decision-making systems that utilize traditional machine learning algorithms. In the context of handwriting tasks, each instance provides seven raw data values: x position, y position, time, pen position indicator, azimuth, altitude, and pressure.

For online handwriting data, feature extraction focuses on capturing dynamic aspects of handwriting. This process involves analyzing temporal characteristics, such as the sequence and timing of pen strokes, and spatial features related to the x and y coordinates. Additionally, it considers the pressure exerted by the pen tip and the angles of azimuth and altitude. Together, these elements create comprehensive feature vectors representing the online handwriting data.

In contrast, offline data, which exists in image form, requires a different approach. Feature extraction for offline data involves image processing and character recognition techniques. The handwritten content is transformed into image features that might include characteristics like line thickness, curvature, and the shape of individual characters. These extracted image features are compiled into feature vectors that encapsulate the distinctive attributes of the offline handwritten content.

Each modality—online and offline—necessitates specialized feature extraction methods to prepare the data for machine learning classification. By tailoring the feature extraction process to the specific type of data, we ensure that the extracted features are meaningful and suitable for building accurate machine learning models.

\subsubsection{Online handwritten features}

From the online data, we have conducted feature extraction across four distinct categories, each serving as a fundamental aspect of handwriting analysis: kinematic, temporal, spatial, and dynamic features. These categories provide a comprehensive understanding of the handwriting process. We followed the same approach as our previous work to extract these features \cite{KUNHOTH2023104715}.
The kinematic features include the horizontal and vertical velocities of writing, which measure the rate of change of the stylus tip's position on the tablet surface in the horizontal and vertical directions with respect to time. The overall velocity of writing combines these components. Horizontal and vertical accelerations represent the rate of change of writing velocity in the respective directions, and the overall acceleration is calculated from these components. Horizontal and vertical jerks measure the rate of change of writing acceleration in the respective directions, with the overall jerk calculated similarly.
Spatial features involve the length of the stroke, measured as the total length of the segment, horizontal length, and vertical length. Additionally, the width and height of the segment represent the maximum horizontal and vertical spans of the stroke, respectively. Temporal features include the duration of the segment, which is the time taken to complete a stroke.
Dynamic features encompass various aspects, such as the pressure exerted by the stylus tip on the writing surface, the altitude (the angle of the stylus pen to the horizontal axis), and the azimuth (the angle of the stylus pen to the vertical axis). Other significant features include the difference between the y-positions of the first and last strokes, the variance of the y-positions of strokes, the number of pen lifts (the number of times the stylus pen tip is lifted from the tablet surface while writing), the count of velocity changes (the number of local extrema in velocity), the count of acceleration changes (the number of local extrema in acceleration), the total duration of writing, and the total length of writing. A total 141 online features were extracted from the online handwritten data.

\subsubsection{Offline handwritten features}

In the context of offline handwriting data analysis, a variety of image processing algorithms can be employed to extract meaningful features. In our approach, we utilized a technique known as "transfer learning via feature extraction." This technique leverages a pre-trained neural network to extract valuable features from a new dataset. Subsequently, these extracted features serve as the foundation for developing a machine learning model, which are then input into supervised learning algorithms such as SVM to facilitate the learning process. Specifically, for our work, we harnessed the power of EfficientNet-B7 \cite{tan2019efficientnet}, a convolutional neural network architecture known for its efficiency and accuracy as a feature extractor.
EfficientNet-B7 is a part of the EfficientNet family, which scales up models in a more balanced way using a compound scaling method. This method uniformly scales the network’s width, depth, and resolution using a set of fixed scaling coefficients. EfficientNet-B7 achieves state-of-the-art accuracy while being computationally efficient. It has been shown to perform exceptionally well on benchmark datasets like ImageNet.
The architecture of EfficientNet-B7 consists of several key components, including MBConv blocks (Mobile Inverted Bottleneck Convolutional blocks), which are designed to optimize both accuracy and efficiency. The network is built with the following considerations:
\begin{itemize}
    \item Compound scaling: EfficientNet-B7 employs compound scaling to balance network depth, width, and resolution, enhancing performance while minimizing computational costs.

    \item MBConv blocks: Each MBConv block includes depthwise separable convolutions, which split the convolution operation into two parts: depthwise convolution and pointwise convolution. This significantly reduces the number of parameters and computational complexity.

    \item Squeeze-and-excitation (SE) blocks: EfficientNet-B7 integrates SE blocks to adaptively recalibrate channel-wise feature responses, further improving model performance.
\end{itemize}

The EfficientNet-B7 architecture includes several stages where the resolution of feature maps is progressively reduced while the number of channels increases, allowing the network to learn more abstract and high-level representations. This structure enhances feature reuse and promotes efficient information flow throughout the network.
In traditional deep CNN models, layers primarily receive input from the immediate preceding layer. However, EfficientNet-B7’s composite operations leverage a more advanced scaling method that ensures balanced growth across all dimensions of the network. This leads to robust feature extraction capabilities.

\subsection{Multi modal data fusion }

\begin{figure*}[h]
	\centerline{\includegraphics[width= 13 cm,height=16 cm]{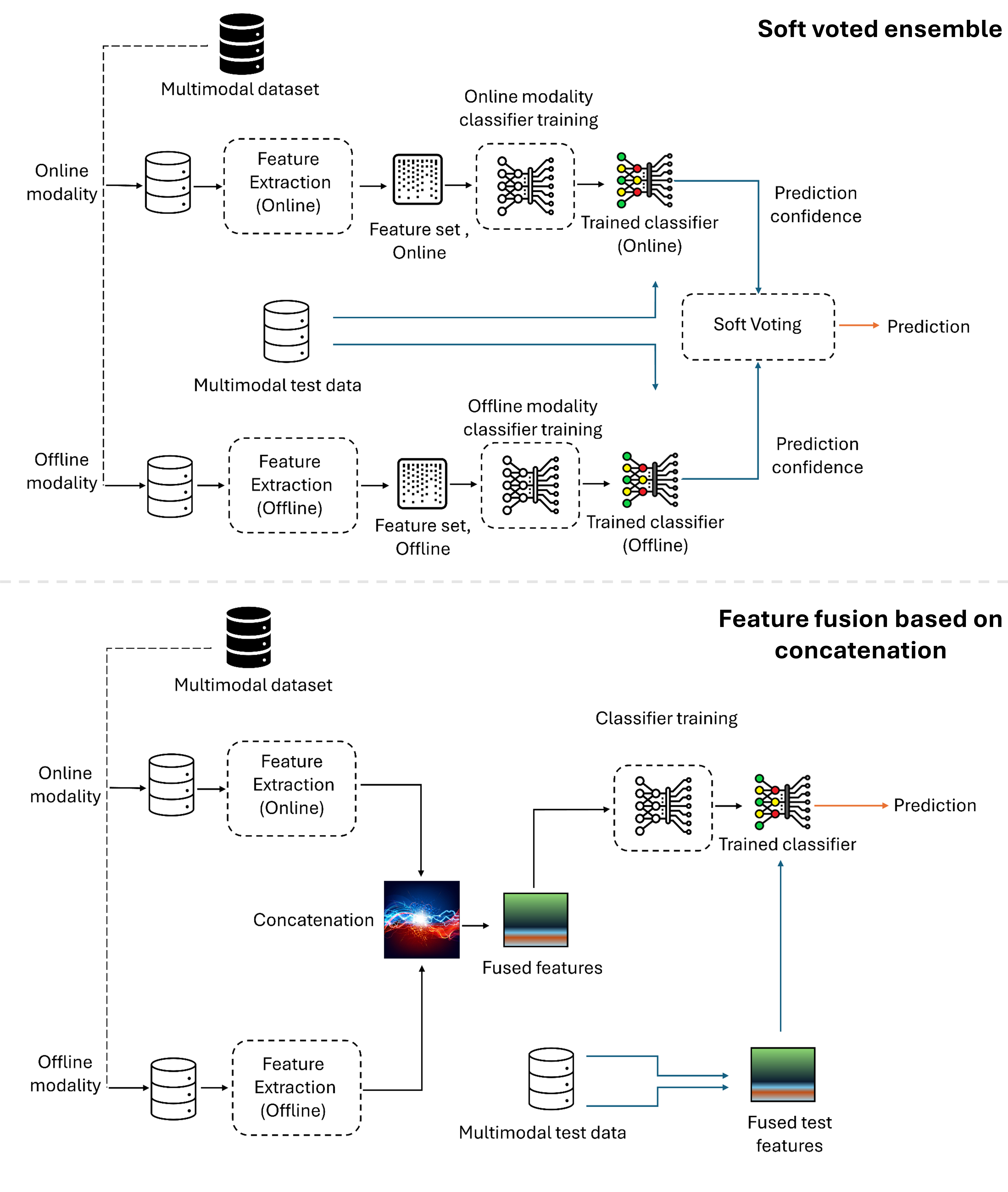}}
	\caption{Overview of implemented traditional multimodal classifier fusion and feature fusion approach for dysgraphia diagnosis}
	\label{fig5}
\end{figure*}

Multimodal learning, a powerful approach in artificial intelligence, integrates information from various sources or modalities, such as text, images, and audio. This method's significant advantage lies in its ability to capture diverse and complementary features, leading to a more comprehensive understanding of the data. By combining different types of information, multimodal learning enhances the model's robustness and performance across a wide range of tasks, making it particularly beneficial in real-world applications.
The implementation of multimodal learning involves extracting relevant features from each modality and merging them to form a unified representation. This can be achieved through methods such as feature concatenation, late fusion (classifier fusion), or early fusion. Effective feature extraction from each modality is essential to maintain the unique characteristics of different data types while ensuring a cohesive integration of information.

In the context of dysgraphia diagnosis, both online handwriting data (which captures dynamic aspects of writing) and offline/image data (which captures static features) are important. Multimodal learning is crucial in this setting because it combines the strengths of both data types, leading to a more accurate and comprehensive assessment of dysgraphia, thereby potentially enhancing diagnostic accuracy.
In our research, we introduced both classifier fusion and feature fusion approaches for multimodal learning in dysgraphia diagnosis. An overview of the traditional classifier fusion approach (based on soft voting) and the feature fusion approach is illustrated in Figure \ref{fig5}.

\subsubsection{Soft voting based ensembling}

Soft voting-based fusion, including strategies like average voting and weighted voting, leverages the combined strengths of multiple classifiers to enhance prediction accuracy. Each classifier contributes its prediction, and these individual outputs are aggregated to form a final decision. Average voting entails averaging class probabilities or confidence scores from each classifier, whereas weighted voting involves assigning different weights to classifiers based on their performance or reliability before combining their probabilities/confidence scores. 

\begin{figure*}[h]
	\centerline{\includegraphics[width= 12 cm,height= 7 cm]{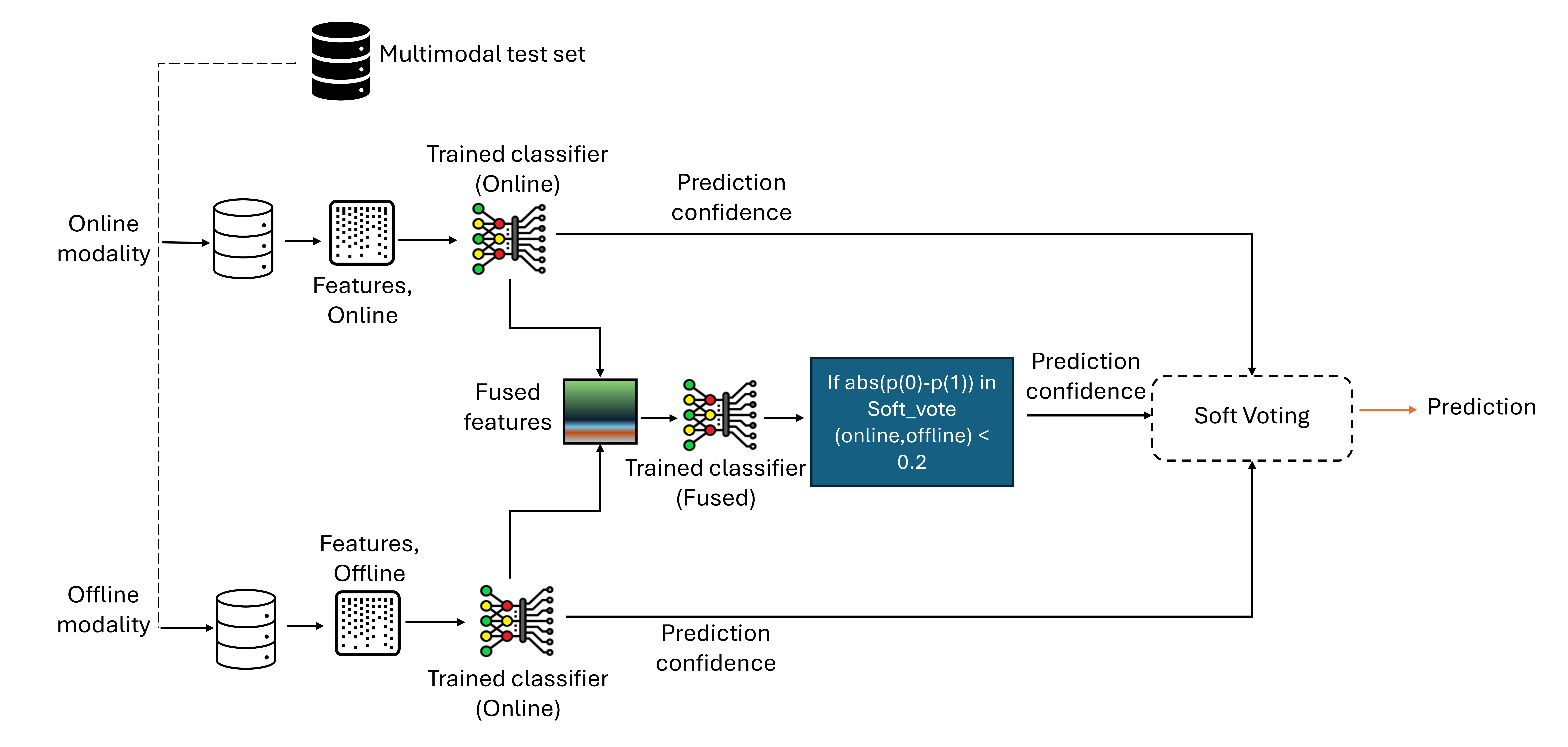}}
	\caption{Overview of proposed novel multimodal classifier ensemble with conditonal feature fusion approach for dysgraphia diagnosis}
	\label{fig4}
\end{figure*}

In this work, we employed average voting for classifier fusion because the performance of the individual modality models (online modality classifier and offline modality classifier) was nearly identical across both datasets (word and pseudo-word). Given this comparable performance, average voting was a suitable method to combine the classifiers' outputs effectively. The softvoting based ensembling employed in this work can be defined as follows: 

Let $Clf_{\text{online}}$ and $Clf_{\text{offline}}$ be the two classifiers trained on online features $X_{\text{online}}$ and offline features $X_{\text{offline}}$, respectively. The features are extracted from the dataset $D = \left(X_{\text{online}}, X_{\text{offline}}, y \right)$, where $y \in \{\text{"TD", "DYG"}\}$, which shows typically developing and dysgraphia categories.
Let $\hat{y}_{\text{online}}$ and $\hat{y}_{\text{offline}}$ represent the predicted probability distribution by classifiers $Clf_{\text{online}}$ and $Clf_{\text{offline}}$ for a given test sample. Then, $\hat{y}_{i}= \left( \hat{y}_{i}^{\text{(TD)}}, \hat{y}_{i}^{\text{(DYG)}} \right)$ for $i = \{\text{online}, \text{offline}\}$.
Here, each $\hat{y}_{i}^{(j)}$ in $\hat{y}_{i}$ is the predicted probability or confidence score of class $j$ by the $i$-th classifier. The $\hat{y}_{i}^{(j)}$ can be fused in different ways. 
In the case of average voting, the average of the probabilities obtained for each class from every classifier is computed.
Let the computed average probability be $
\hat{y}_{\text{final}}= \left(\hat{y}^{\text{(TD)}}, \hat{y}^{\text{(DYG)}}\right)
$
and $\hat{y}^{(j)}$ in $\hat{y}_{\text{final}}$ is defined as 
$
\hat{y}^{(j)} = \frac{1}{2} \left(\hat{y}_{\text{online}}^{(j)} + \hat{y}_{\text{offline}}^{(j)}\right)
$.
Then the fused prediction is defined as $
\hat{y} = \arg\max(\hat{y}_{\text{final}})$.

\subsubsection{Feature fusion via concatenation}

In this work, we employed feature fusion by concatenating features from online and offline handwritten data. This approach involves combining features extracted from both modalities into a single feature vector, which is then used to train a machine learning algorithm. During testing, features are similarly extracted from the test data, concatenated, and used for prediction.

Let $X_{\text{online}}$ and $X_{\text{offline}}$ represent the feature sets extracted from the online and offline handwritten data, respectively. For a given sample, the feature vectors can be $\mathbf{x}_{\text{online}} \in \mathbb{R}^{m} \quad \text{and} \quad \mathbf{x}_{\text{offline}} \in \mathbb{R}^{n}
$, where $m$ and $n$ are the dimensions of the online and offline feature vectors, respectively.
The feature fusion via concatenation is performed by linearly concatenating these feature vectors to form a single feature vector $(\mathbf{x}_{\text{fused}} = [\mathbf{x}_{\text{online}} \, | \, \mathbf{x}_{\text{offline}}]$ where $\mathbf{x}_{\text{fused}} \in \mathbb{R}^{m+n})$.
This concatenated feature vector $\mathbf{x}_{\text{fused}}$ is then used as input to a machine learning algorithm for training.
Given a dataset $D = \{(X_{\text{online}}^i, X_{\text{offline}}^i, y^i)\}_{i=1}^{N}$, where $N$ is the number of samples, and $y^i \in \{\text{TD, DYG}\}$ represents the class labels (TD for typically developing and DYG for dysgraphia), the concatenated feature vector for all samples is $\mathbf{X}_{\text{fused}} = [\mathbf{x}_{\text{fused}}^1, \mathbf{x}_{\text{fused}}^2, \ldots, \mathbf{x}_{\text{fused}}^N]$ where $\mathbf{X}_{\text{fused}} \in \mathbb{R}^{N \times (m+n)}$.
A machine learning algorithm is then trained on $\mathbf{X}_{\text{fused}}$ and the corresponding labels $y = [y^1, y^2, \ldots, y^N]$.

\textbf{Testing Phase}
During testing, features are extracted from the test sample, concatenated in the same manner, and used for prediction. For a test sample, let $\mathbf{x}_{\text{online}}^{\text{test}}$ and $\mathbf{x}_{\text{offline}}^{\text{test}}$ be the extracted features. $\mathbf{x}_{\text{fused}}^{\text{test}} = [\mathbf{x}_{\text{online}}^{\text{test}} \, | \, \mathbf{x}_{\text{offline}}^{\text{test}}]$ is the fused test feature vector.
This fused feature vector $\mathbf{x}_{\text{fused}}^{\text{test}}$ is then passed to the trained machine learning model to obtain the final prediction.

\subsubsection{Ensemble with conditional feature fusion}

Ensemble with conditional feature fusion/classifier fusion with conditional feature fusion is a novel technique employed in our dysgraphia diagnosis framework to enhance the reliability of decision-making. It addresses situations where the confidence of the initial ensemble prediction falls below a predefined threshold, indicating uncertainty in the classification outcome. In such cases, instead of relying solely on the initial ensemble, we perform an additional round of ensemble voting. The workflow of proposed ensemble with conditional feature fusion approach is provided in Figure 4.

In our approach, we first train single-modality classifiers, $Clf_{\text{online}}$ and $Clf_{\text{offline}}$, on online and offline handwriting data, respectively. These classifiers provide probability distributions for dysgraphia and typically developing classes. Next, we employ average voting ensembling to combine the outputs of $Clf_{\text{online}}$ and $Clf_{\text{offline}}$, yielding an initial ensemble prediction.

To further leverage the complementary information captured by the two modalities, we introduce feature fusion by concatenating the online and offline modality features. This fusion process generates a combined feature vector, which is used to train another classifier, denoted as $Clf_{\text{fused}}$. The intuition behind feature fusion is to provide the classifier with richer information that encompasses both modalities, potentially enhancing its discriminative power.
However, to ensure robustness in decision-making, we incorporate a conditional feature fusion mechanism.

\begin{algorithm} 
\caption{Multimodal Classifier Fusion with Conditional Feature Fusion}\label{multimodal_algorithm}

\begin{algorithmic}[1]
\Require Classifier trained on online  features $Clf_{\text{online}}$, classifier trained on offline  features $Clf_{\text{offline}}$, classifier trained on fused features $Clf_{\text{fusion}}$, test data $D_{test}=[D_{test}^{(online)},D_{test}^{(offline)}]$, threshold $\tau$,
\Ensure Final predicted label $\hat{y}_{\text{final}}$
\Algphase{Classifier prediction}
\Procedure{MultimodalClassifierFusion}{$Clf_{\text{online}}$, $Clf_{\text{offline}}$, $Clf_{\text{fusion}}$, $D_{\text{test}}$, $\tau$} 

\State \textcolor{green}{// Predict probabilities for single modality}
    \State $P_{\text{online}} \gets Clf_{\text{online}}(D_{test}^{(online)})$ 
    \State $P_{\text{offline}} \gets Clf_{\text{offline}}(D_{test}^{(offline)})$

    \State \textcolor{green}{// Soft voting ensembling of single modality classifier}
    \State $P_{\text{ensemble}} \gets \text{SoftVote}(P_{\text{online}}, P_{\text{offline}})$
    
    \If{$\max(P_{\text{ensemble}}) < \tau$} \State \textcolor{green}{// Conditional Feature Fusion}
        
        \State $P_{\text{fusion}} \gets Clf_{\text{fusion}}(D_{test}^{(online)} \oplus D_{test}^{(offline)})$
        \State \textcolor{green}{// Perform another round of soft voting with individual classifiers} 
        \State $P_{\text{soft\_vote}} \gets \text{SoftVote}(P_{\text{online}}, P_{\text{offline}}, P_{\text{fusion}})$
        \State $\hat{y}_{\text{final}} \gets \arg\max(P_{\text{soft\_vote}})$
    \Else
        \State $\hat{y}_{\text{final}} \gets \arg\max(P_{\text{ensemble}})$
    \EndIf
    
    \State \textbf{Return} $\hat{y}_{\text{final}}$
    
\EndProcedure
\Algphase{Soft voting}
\Procedure{SoftVote}{$P$}
    \State $P_{\text{ensemble}} \gets \frac{1}{len(P)} \sum_{i=1}^{len(P)} P_{i}$
    \State \textbf{Return} $P_{\text{ensemble}}$
\EndProcedure

\end{algorithmic}

\end{algorithm}

After obtaining the initial ensemble prediction, we evaluate the confidence scores for dysgraphia and typically developing classes. If the confidence score for any class falls below a predefined threshold (e.g., 0.2), indicating uncertainty in the decision, we perform another round of soft voting ensemble. This time, we include the outputs of $Clf_{\text{online}}$, $Clf_{\text{offline}}$, and $Clf_{\text{fused}}$ in the ensemble. The final decision is made based on the combined ensemble output, thereby mitigating potential errors caused by low confidence scores.
The importance of conditional feature fusion lies in its ability to mitigate potential errors caused by low confidence scores in the initial ensemble prediction. By incorporating this conditional mechanism, we ensure that the final decision is based on a more comprehensive assessment of the available information. This approach enhances the robustness and reliability of our dysgraphia diagnosis model, ultimately leading to improved diagnostic accuracy.
In practical terms, conditional feature fusion provides a safety net against unreliable predictions by triggering a secondary evaluation process when confidence in the initial ensemble is insufficient. This mechanism adds a layer of adaptability to our diagnostic framework, allowing it to dynamically adjust its decision-making strategy based on the quality of available evidence. As a result, our method can effectively handle cases where individual modality classifiers or initial ensemble predictions may be less reliable, ensuring more accurate and consistent dysgraphia diagnosis outcomes. The conditional feature fusion process is detailed in Algorithm \ref{multimodal_algorithm}.

\section{Evaluation and Results}

Assessing and evaluating the efficacy of proposed methodologies is crucial for determining their ability to effectively tackle the problem at hand. In this research, a series of experiments were conducted to scrutinize and analyze the performance of the proposed techniques. All experimental procedures were implemented using the Python programming language. The training and evaluation phases were executed on a computing system equipped with an Intel(R) Core(TM) i7-7820HK CPU, operating at a frequency of 2.90 GHz (2901 MHz) with four processing cores, and an Nvidia GTX 1060 graphics processing unit (GPU). The well-established SciKit library was employed for the implementation of conventional machine learning algorithms, while the TensorFlow framework was utilized for the realization of feature extraction using a deep convolutional neural network (CNN).

To ensure a comprehensive performance analysis of the proposed methods, a diverse set of evaluation metrics were considered. The metrics employed in this study include accuracy, precision, and recall scores. While accuracy is a commonly used metric for classification tasks, precision, and recall are particularly effective for handling datasets with class imbalance.

In order to conduct a thorough investigation of classifier performance and mitigate the potential impact of random selection bias, a stratified group ten-fold cross-validation approach was employed for both training and evaluation. This methodology ensures that each cross-validation fold maintains a class label distribution that is consistent with the original dataset. Furthermore, hyperparameter tuning was performed using a grid search technique to identify the optimal hyperparameter configuration for each classifier. This approach allows for the exploration of a wide range of hyperparameter combinations, enabling the selection of the most suitable values for maximizing classifier performance.

\subsection{Individual modality analysis}

Table 2 presents a comparative analysis of the performance of two machine learning algorithms, SVM and XGboost, in diagnosing dysgraphia using online and offline modalities of handwritten data. The evaluation metrics used to assess the performance of these algorithms include accuracy, precision, and recall.

\begin{table*}[]
\begin{tabular}{llcccccc}
\hline
                           &                        & \multicolumn{3}{c}{SVM}                  & \multicolumn{3}{c}{XGboost}              \\ \cline{3-8} 
\multirow{-2}{*}{Modality} & \multirow{-2}{*}{Data} & Accuracy & Precision & Recall & Accuracy & Precision & Recall \\ \hline \hline
                           & Dword                  & 72.1   & 74.0   & 60.6   & 72.6   & 73.7   & 68.8   \\ \cline{2-8} 
                           & Word                   & 72.6   & 71.8   & 81.0   & 72.8   & 76.5   & 69.7  \\ \cline{2-8} 
\multirow{-3}{*}{Online}   & Pword                  & 76.2   & 79.0   & 70.0   & 71.1   & 74.3   & 65.2   \\ \hline
                           & Dword                  & 66.0   & 64.3   & 62.0   & 66.0   & 62.7  & 59.4   \\ \cline{2-8} 
                           & Word                   & 74.6   & 81.7   & 70.0   & 76.78   & 80.86   & 76.6   \\ \cline{2-8} 
\multirow{-3}{*}{Offline}  & Pword                  & 76.3   & 79.3   & 73 .2  & 72.9   & 75.4   & 69.6   \\ \hline
\end{tabular}
\caption{Performance comparison of SVM and XGboost for online and offline modalities of handwritten data. Here Word, Pword/psuedoword and Dword/diffult word represents words 'leto','lamoken' and 'hra\v{c}k'arstvo' respectively.}

\end{table*}

In the online modality, both SVM and XGboost demonstrate similar performance across different data types (Dword, Word, and Pword). For Dword data, SVM achieves an accuracy of 72.1\%, precision of 74.0\%, and recall of 60.6\%, while XGboost shows slightly better results with an accuracy of 72.6\%, precision of 73.7\%, and recall of 68.8\%. For Word data, SVM and XGboost exhibit comparable accuracies of 72.6\% and 72.8\%, respectively. However, XGboost achieves higher precision (76.5\%) compared to SVM (71.8\%), while SVM has a higher recall (81.0\%) compared to XGboost (69.7\%). In the case of Pword data, SVM outperforms XGboost with an accuracy of 76.2\%, precision of 79.0\%, and recall of 70.0\%, compared to XGboost's accuracy of 71.1\%, precision of 74.3\%, and recall of 65.2\%.

Moving to the offline modality, both algorithms show similar accuracies for Dword data (66.0\%). However, SVM has slightly higher precision (64.3\%) and recall (62.0\%) compared to XGboost (precision: 62.7\%, recall: 59.4\%). For Word data, XGboost demonstrates better performance with an accuracy of 76.78\%, precision of 80.86\%, and recall of 76.6\%, while SVM achieves an accuracy of 74.6\%, precision of 81.7\%, and recall of 70.0\%. In the case of Pword data, SVM exhibits higher accuracy (76.3\%), precision (79.3\%), and recall (73.2\%) compared to XGboost (accuracy: 72.9\%, precision: 75.4\%, recall: 69.6\%).
In the online modality, SVM demonstrates slightly better performance for Pword data, while XGboost shows marginally better results for Dword data. In the offline modality, XGboost outperforms SVM for Word data, while SVM exhibits better performance for pword data.It is important to note that the differences in performance between the two algorithms are relatively small.

Upon further analysis of the results, it is evident that both SVM and XGboost demonstrate relatively low performance in diagnosing dysgraphia using Dword data, particularly in the offline modality. This can be attributed to the inherent difficulty in distinguishing between the handwriting patterns of typically developing individuals and those with dysgraphia when examining offline handwritten samples of difficult words.

In the offline modality, the accuracy for Dword data is 66.0\% for both SVM and XGboost, which is considerably lower compared to the accuracies achieved for word and pword data. The precision and recall values for Dword data in the offline modality are also lower, with SVM achieving a precision of 64.3\% and a recall of 62.0\%, while XGboost shows a precision of 62.7\% and a recall of 59.4\%. These results highlight the challenges associated with accurately identifying dysgraphia based solely on offline handwriting samples of difficult words.
The difficulty in distinguishing between typically developing handwriting and dysgraphic handwriting in offline Dword data can be attributed to several factors. Firstly, the static nature of offline handwriting samples limits the available information regarding the temporal and kinematic aspects of the writing process, which are crucial in detecting the subtle differences between normal and dysgraphic handwriting. Secondly, the complexity and variability of difficult words may obscure the characteristic features of dysgraphia, making it harder for the algorithms to capture the discriminative patterns.
In contrast, the online modality provides additional temporal and kinematic information, such as stroke order, velocity, and pressure, which can aid in the identification of dysgraphic handwriting patterns. This is reflected in the relatively higher accuracies, precisions, and recalls achieved by both SVM and XGboost for Dword data in the online modality compared to the offline modality.

\subsection{Multimodality analysis}
Considering the poor performance of both algorithms in diagnosing dysgraphia using offline Dword data, we have decided to focus our multimodality analysis on Word and Pword data only. By concentrating on these data types, which yield better results in both online and offline modalities, we aim to leverage the complementary information provided by the two modalities to enhance the accuracy and reliability of dysgraphia diagnosis.

\begin{table*}[b]
\begin{tabular}{llcccccc}
\hline
& & \multicolumn{3}{c}{SVM} & \multicolumn{3}{c}{XGboost} \\ \cline{3-8} 
\multirow{-2}{*}{Multimodality fusion method} & \multirow{-2}{*}{Data} & Accuracy & Precision & Recall & Accuracy & Precision & Recall \\ \hline \hline
& Word & 77.9 & 84.3 & 73.1 & 80.9 & 84.23 & 80.5 \\ \cline{2-8} 
\multirow{-2}{*}{Feature fusion} & Pword & 83.4 & 84.7 & 81.6 & 80.5 & 84.1 & 75.6 \\ \hline
& Word & 77.4 & 81.67 & 77.5 & 78.1 & 80.05 & 77.9 \\ \cline{2-8} 
\multirow{-2}{*}{Soft voting ensemble} & Pword & 85.6 & 84.4 & 88.6 & 76.6 & 79.7 & 73.9 \\ \hline
\end{tabular}

\caption{Performance comparison of SVM and XGboost using feature fusion and soft voted ensemble methods. Here Word and Pword/psuedoword represents words 'leto', and'lamoken' respectively.}
\end{table*}
Table 3 presents the performance of two multimodality fusion methods, feature fusion and soft voting ensemble, for diagnosing dysgraphia using SVM and XGboost classifiers. The evaluation metrics used are accuracy, precision, and recall. The analysis is conducted on two types of data: Word and Pword.

In the feature fusion method, the features from both modalities are concatenated and then used to train and test the classifiers. For Word data, XGboost achieves a higher accuracy (80.9\%), precision (84.23\%), and recall (80.5\%) compared to SVM (accuracy: 77.9\%, precision: 84.3\%, recall: 73.1\%). However, for Pword data, SVM outperforms XGboost with an accuracy of 83.4\%, precision of 84.7\%, and recall of 81.6\%, while XGboost obtains an accuracy of 80.5\%, precision of 84.1\%, and recall of 75.6\%.

In the soft voting ensemble method, each classifier is trained separately on the individual modalities, and the final prediction is made by averaging the confidence of the classifiers during testing. For Word data, XGboost shows slightly better performance than SVM, with an accuracy of 78.1\%, precision of 80.05\%, and recall of 77.9\%, compared to SVM's accuracy of 77.4\%, precision of 81.67\%, and recall of 77.5\%. However, for Pword data, SVM significantly outperforms XGboost, achieving an accuracy of 85.6\%, precision of 84.4\%, and recall of 88.6\%, while XGboost obtains an accuracy of 76.6\%, precision of 79.7\%, and recall of 73.9\%.

Overall, the results suggest that the performance of the multimodality fusion methods varies depending on the type of data and the classifier used. For Word data, XGboost generally performs better than SVM in both feature fusion and soft voting ensemble methods. However, for Pword data, SVM consistently outperforms XGboost in both fusion methods.

It is worth noting that the soft voting ensemble method with SVM achieves the highest accuracy (85.6\%), precision (84.4\%), and recall (88.6\%) for Pword data among all the combinations of fusion methods, data types, and classifiers. This indicates that the soft voting ensemble approach, which leverages the complementary information from multiple modalities by training classifiers separately and combining their predictions, can effectively improve the diagnosis of dysgraphia, particularly for Pword data.
It is clearly evident from the results that the multimodal approach has significantly improved the performance of dysgraphia diagnosis compared to using single modalities. However, it is important to note that multimodal analysis comes with increased computational overhead due to the larger number of features and classifiers involved. To further enhance the diagnosis performance without significantly increasing the computational burden, we propose a novel approach called ensemble with conditional feature fusion.

In our proposed method, we initially perform soft voting between the online and offline classifiers. If the confidence score difference between the positive and negative classes is less than 0.2 in pseudoword data and 0.15 in word data ( value is selected after expereimenting with set of threshold values), indicating a level of uncertainty in the prediction, we proceed to perform another round of soft voting. This additional voting includes classifiers trained on online features, offline features, and fused features, leveraging the complementary information from all available modalities.
The results of our proposed ensemble with conditional feature fusion approach are presented in Table 4. This novel method aims to strike a balance between improving the diagnosis performance and maintaining computational efficiency by conditionally incorporating feature fusion based on the confidence of the initial soft voting prediction.

\begin{table*}[]
\begin{tabular}{lcccccc}
\hline
& \multicolumn{3}{c}{SVM} & \multicolumn{3}{c}{XGboost} \\ \cline{2-7} 
\multirow{-2}{*}{Data} & Accuracy & Precision & Recall & Accuracy & Precision & Recall \\ \hline \hline
word & 80.0 & 84.6 & 78.3 & 78.8 & 82.5 & 77.1 \\ \hline
pword & 88.8 & 88.8 & 90.0 & 80.0 & 83.0 & 78.0 \\ \hline
\end{tabular}
\caption{Performance of novel conditional ensemble with feature fusion approach in Word and Pword data. Here Word and Pword/psuedoword represents words 'leto', and'lamoken' respectively. }
\end{table*}

For Word data, the SVM classifier achieves an accuracy of 80 \%, precision of 84.6\%, and recall of 77.3\%. In comparison, the XGboost classifier obtains an accuracy of 78.8\%, precision of 82.5\%, and recall of 77.1\%. The SVM classifier slightly outperforms XGboost in all three metrics for word data, indicating its effectiveness in capturing the discriminative patterns for dysgraphia diagnosis.
Moving on to Pword data, the SVM classifier demonstrates exceptional performance with an accuracy of 88.8\%, precision of 88.8\%, and recall of 90.0\%. This suggests that the proposed ensemble method is highly effective in correctly identifying individuals with dysgraphia based on their Pword handwriting samples. On the other hand, the XGboost classifier achieves an accuracy of 80.0\%, precision of 83.0\%, and recall of 78.0\% for pword data, which is comparatively lower than SVM but still indicates a strong performance.
The proposed ensemble method with conditional feature fusion leverages the strengths of both online and offline modalities by initially performing soft voting between their respective classifiers. In cases where the confidence score difference between the positive and negative classes is less than 0.2, indicating a level of uncertainty in the prediction, the method incorporates an additional round of soft voting that includes classifiers trained on fused features. This conditional feature fusion approach allows for the selective integration of complementary information from multiple modalities, enhancing the diagnosis performance without significantly increasing the computational overhead.
The results demonstrate that the proposed method is particularly effective for pword data, where the SVM classifier achieves remarkably high accuracy, precision, and recall values. This suggests that the ensemble method successfully captures the subtle characteristics of dysgraphic handwriting in Pword samples, enabling accurate diagnosis. The performance on Word data is also commendable, with both SVM and XGboost classifiers achieving accuracy values close to 80\% and precision values above 80\%.
One notable observation is the consistent superiority of the SVM classifier over XGboost in both Word and Pword data. This indicates that SVM's ability to find optimal decision boundaries in high-dimensional feature spaces is particularly suitable for the task of dysgraphia diagnosis using the proposed ensemble method.

\begin{figure*}[b]
	\centerline{\includegraphics[width= 16 cm,height=5 cm]{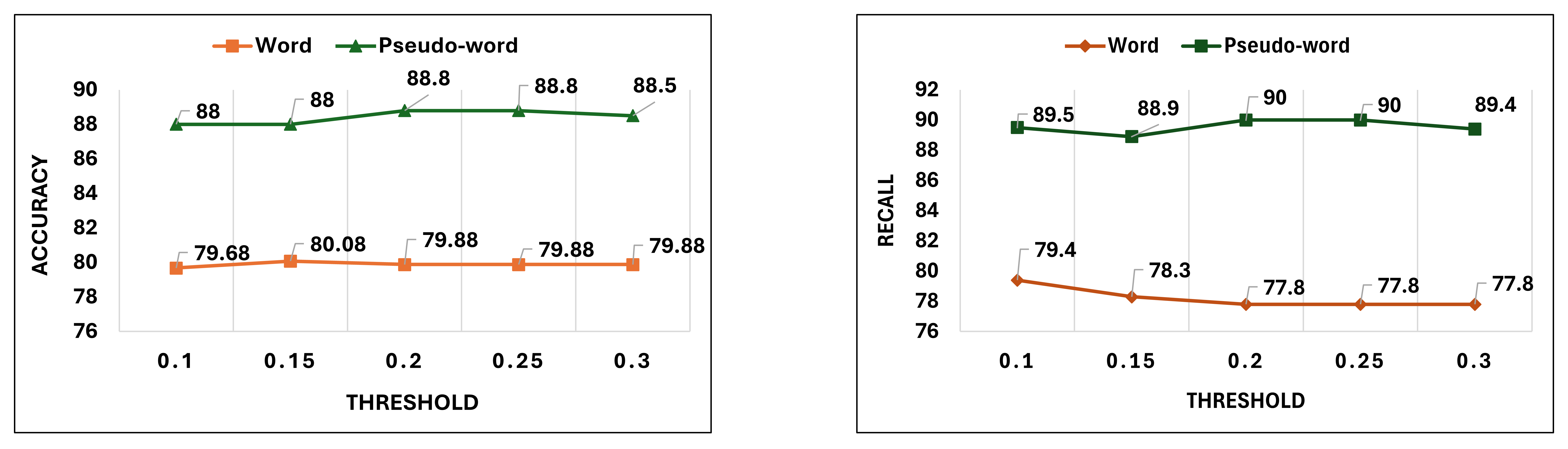}}
	\caption{Effect of confidence score threshold value in multimodal ensemble with conditional feature fusion method for  word and pseudo word data. }
	\label{fig3}
\end{figure*}

To further analyze the impact of the confidence score threshold on the performance of the proposed ensemble method with conditional feature fusion, we conducted experiments with different threshold values. While the primary results presented in the study were based on a threshold of 0.2, it is essential to explore how varying this threshold affects the diagnosis accuracy, precision, and recall.

We selected a range of threshold values, including 0.1, 0.15, 0.25, and 0.3, in addition to the original 0.2 threshold. For each threshold value, we evaluated the performance of the ensemble method on both Word and Pword data using SVM classifier. This analysis allows us to assess the sensitivity of the proposed method to the confidence score threshold and identify any potential improvements or trade-offs. Figure 5 displays the accuracy and recall values obtained for different threshold values in ensemble with conditional feature fusion ( in SVM classifier) in word and psuedoword data.

For Word data, the accuracy of the ensemble method remains relatively stable across the explored threshold range, with minor fluctuations. The highest accuracy of 80.0\% is achieved at a threshold of 0.15, while the accuracy at other thresholds ranges from 79.6\% to 79.8\%. This suggests that the ensemble method is not highly sensitive to the confidence score threshold for Word data, and a threshold of 0.15 may be optimal for achieving the best accuracy.

Similarly, recall value for Word data exhibit stability across different thresholds.The recall values range from 77.8\% to 79.4\%, with the highest recall achieved at a threshold of 0.10. These results indicate that the ensemble method maintains consistent performance in terms of accuracy and recall for Word data, with a threshold of 0.15 offering a good balance between the two metrics.

Moving on to Pword data, we observe a similar trend of stable performance across different confidence score thresholds. The accuracy of the ensemble method ranges from 88.0\% to 88.8\%, with the highest accuracy achieved at thresholds of 0.20 and 0.25. This suggests that the ensemble method is robust and maintains high accuracy for Pword data, even when the confidence score threshold is varied.
The recall values exhibit a slightly wider range, from 88.9\% to 90.0\%, with the highest recall achieved at thresholds of 0.20 and 0.25. These results demonstrate that the ensemble method performs exceptionally well on Pword data, maintaining high accuracy and recall across different thresholds. The thresholds of 0.20 and 0.25 seem to be optimal for achieving the best overall performance on Pword data.

The analysis of different confidence score thresholds reveals that the proposed ensemble method with conditional feature fusion is relatively robust and maintains consistent performance across the explored threshold range for both Word and Pword data. The method is not highly sensitive to the specific threshold value, indicating its ability to effectively leverage the complementary information from multiple modalities for accurate dysgraphia diagnosis.

However, it is important to note that the optimal threshold value may vary depending on the specific dataset and the desired trade-off between accuracy and recall. For Word data, a threshold of 0.15 seems to offer the best balance, while for Pword data, thresholds of 0.20 and 0.25 yield the highest performance measures.

\subsection{Comparison with state of the art methods}

The proposed methods' effectiveness is highlighted by comparing their performance with state of the art dysgraphia diagnosis techniques evalauted on the same dataset. Table 5 illustrates this performance comparison, with the proposed methods emphasized in bold.

\begin{table}[]
	\centering
	\label{tab6}
	\caption{Comparison with state-of-the-art methods. Here FF is feature fusion, SVE is soft vote ensemble, ECFF is ensemble with conditional feature fusion}
	\begin{tabular}{p{3cm}p{2.3cm}c}
		\hline
		Methods                     & Data type & Accuracy \\ \hline \hline
		AdaBoost \cite{Drotar2020}                    & Online    & 79.5\%   \\ \hline
		AdaBoost \cite{KUNHOTH2023104715}                         & Online     & 80.8\%   \\ \hline
        CNN \cite{skunda2022method} & Online     & 79.7\%   \\ \hline
		\textbf{SVM with FF} & \textbf{Multimodal }     &  \textbf{83.4\% }  \\ \hline
        \textbf{SVM with SVE} & \textbf{Multimodal }     &  \textbf{85.6\% }  \\ \hline
        \textbf{SVM with ECFF} & \textbf{Multmodal }     &  \textbf{88.8\% }  \\ \hline

	\end{tabular}
\end{table}

The Table 5 presents a comparison of the proposed multimodal methods with state-of-the-art techniques for dysgraphia diagnosis. The proposed methods, highlighted in bold, include SVM with feature fusion (FF), SVM with soft vote ensemble (SVE), and SVM with ensemble and conditional feature fusion (ECFF). These methods are compared against AdaBoost \cite{Drotar2020}, AdaBoost \cite{KUNHOTH2023104715}, and CNN \cite{skunda2022method}, which utilize only online data.
The state-of-the-art methods using online data achieve accuracies ranging from 79.5\% to 80.8\%. AdaBoost \cite{Drotar2020} obtains an accuracy of 79.5\%, while AdaBoost \cite{KUNHOTH2023104715} slightly improves upon it with an accuracy of 80.8\%. The CNN method \cite{skunda2022method} achieves an accuracy of 79.7\%, which is comparable to the AdaBoost methods.
In contrast, the proposed multimodal methods demonstrate superior performance. SVM with feature fusion achieves an accuracy of 83.4\%, outperforming all the state-of-the-art methods that rely solely on online data. This indicates that the integration of features from both online and offline modalities enhances the discriminative power of the classifier, leading to improved diagnosis accuracy.
Furthermore, SVM with soft vote ensemble (SVE) further improves the accuracy to 85.6\%. This method leverages the complementary information from online and offline modalities by training separate classifiers and combining their predictions through soft voting. The increased accuracy suggests that the ensemble approach effectively captures the diverse characteristics of dysgraphia from multiple modalities, resulting in more reliable and accurate diagnosis.
The most notable performance is achieved by SVM with ensemble and conditional feature fusion (ECFF), which obtains an accuracy of 88.8\%. This method intelligently combines the predictions from online and offline classifiers and selectively incorporates feature fusion when the confidence score difference between the positive and negative classes is below a threshold. By conditionally integrating the fused features, ECFF strikes a balance between leveraging the complementary information from multiple modalities and maintaining computational efficiency. The high accuracy achieved by ECFF demonstrates its effectiveness in accurately diagnosing dysgraphia, even when analyzing a single instance of Pord data.
It is important to highlight that the proposed multimodal methods, particularly SVM with ECFF, achieve state-of-the-art performance while analyzing only a single instance of pword data. This is in contrast to the other methods that utilize multiple instances or longer sequences of online data. The ability to accurately diagnose dysgraphia using a single pword sample showcases the efficiency and practicality of the proposed methods, as they require minimal data collection and processing overhead.

In conclusion, the comparison with state-of-the-art methods demonstrates the superiority of the proposed multimodal approaches for dysgraphia diagnosis. The integration of online and offline modalities through feature fusion, soft vote ensemble, and ensemble with conditional feature fusion significantly improves the accuracy of dysgraphia diagnosis. The SVM with ECFF method, in particular, achieves the highest accuracy of 88.8\% while analyzing only a single instance of Pword data, outperforming the state-of-the-art methods that rely solely on online data. 
\section{Discussion}
In this study, we have presented a novel approach for dysgraphia diagnosis using machine learning algorithms on a multimodal dataset. Our work addresses the limitations of existing research, which primarily focuses on single modality data, either online or offline handwriting, with a majority of studies concentrating on online data. To bridge this gap, we created a new dataset by transforming an existing online handwritten dataset through rasterization, generating corresponding offline handwriting images. This dataset encompasses various handwriting tasks, including letter writing, syllable writing, word writing, pseudoword writing, difficult word writing, and sentence writing.
One of the key objectives of our work is to accurately detect dysgraphia from a single instance of a word, focusing on three word types: word, pseudoword, and difficult word. We trained SVM and XGBoost classifiers separately on online and offline features and analyzed the results. Our findings indicate that both classifiers demonstrate comparable performance across different word types in the online modality. However, their performance was relatively lower for difficult word data in the offline modality, with accuracies around 66\%, highlighting the challenges in distinguishing between normal and dysgraphic handwriting in complex offline samples. This poor performance can be attributed to the visual similarity between normal and dysgraphic handwriting for difficult words, making it harder for the classifiers to capture discriminative features in the offline modality.
To leverage the complementary information from both modalities, we implemented two multimodal approaches: feature fusion and soft-voted ensembling. These approaches yielded improved performance compared to the single modality experiments. The results varied depending on the word type and classifier used, with XGBoost generally outperforming SVM for word data and SVM surpassing XGBoost for pseudoword data in both fusion methods. Notably, the soft-voted ensembling approach with SVM achieved the highest accuracy of 85.6\%, precision of 84.4\%, and recall of 88.6\% for pseudoword data among all combinations of fusion methods, data types, and classifiers. This represents a significant improvement of approximately 9-10\% in accuracy compared to the single modality experiments.
Furthermore, we proposed a novel approach called ensemble with conditional feature fusion, which intelligently combines the predictions from online and offline classifiers and selectively incorporates feature fusion when the confidence score difference between the positive and negative classes falls below a threshold. This approach aims to strike a balance between improving diagnosis performance and maintaining computational efficiency. Our results demonstrate that the proposed method is particularly effective for pseudoword data, where the SVM classifier achieves remarkably high accuracy of 88.8\%, precision of 88.8\%, and recall of 90.0\%. This represents a substantial improvement of around 12-14\% in accuracy compared to the single modality experiments and a 3-4\% improvement over the soft-voted ensembling approach.
The superior performance of our novel approach can be attributed to its ability to selectively incorporate feature fusion based on the confidence scores of the initial soft voting. By conditionally combining the features from both modalities when there is uncertainty in the predictions, the ensemble with conditional feature fusion method effectively leverages the complementary information to enhance the diagnosis accuracy. This approach intelligently utilizes the strengths of both modalities while maintaining computational efficiency, leading to improved performance compared to traditional multimodal fusion methods.
The superiority of our proposed multimodal approaches is further highlighted by comparing their performance with state-of-the-art dysgraphia diagnosis techniques evaluated on the same dataset. The SVM with ensemble and conditional feature fusion (ECFF) method achieves the highest accuracy of 88.8\% while analyzing only a single instance of pseudoword data, outperforming the state-of-the-art methods that rely solely on online data by a margin of 8-9\%.
Our work makes several significant contributions to the field of dysgraphia diagnosis. Firstly, we have created a new multimodal dataset by transforming an existing online handwritten dataset, enabling the exploration of both online and offline modalities. Secondly, we have demonstrated the effectiveness of combining online and offline features through various multimodal approaches, particularly the ensemble with conditional feature fusion method, which yields substantial performance improvements over single modality experiments and state-of-the-art methods. Thirdly, our proposed method achieves state-of-the-art performance while requiring only a single instance of pseudoword data, showcasing its efficiency and practicality in real-world scenarios.

Despite the promising results, our study has some limitations. The dataset used in this research is relatively small, and future work should focus on evaluating the proposed methods on larger and more diverse datasets. Additionally, the interpretability of the machine learning models could be further investigated to gain insights into the specific features and patterns that contribute to the accurate diagnosis of dysgraphia.

\section{Conclusion}

Dysgraphia is a learning disorder that affects an individual's ability to write legibly and efficiently, which can have a significant impact on their academic performance and overall quality of life. Early and accurate diagnosis of dysgraphia is crucial for providing timely interventions and support to those affected. In this study, we have tackled the challenge of dysgraphia diagnosis by proposing a novel multimodal approach that leverages machine learning algorithms and combines information from both online and offline handwriting data. Our research has made significant strides in advancing the field of dysgraphia diagnosis by addressing the limitations of previous studies that primarily focused on single modality data. By creating a new multimodal dataset from existing online dataset and exploring various fusion techniques, including feature fusion, soft-voted ensembling, and our innovative ensemble with conditional feature fusion method, we have demonstrated the immense potential of multimodal learning in enhancing the accuracy and efficiency of diagnostic tools.

The results of our study are highly promising, with our proposed ensemble with conditional feature fusion method achieving state-of-the-art performance (88.8\% accuracy and 90\% recall), surpassing existing methods by a considerable margin. The implications of our work are far-reaching, as it paves the way for the development of more effective and accessible diagnostic tools for dysgraphia. By requiring only a single instance of multimodal word data (word or pseudoword), our approach demonstrates its practicality and potential for real-world applications, making it easier for professionals to accurately identify individuals with dysgraphia and provide them with the necessary support.

\section*{Acknowledgement}

This study was funded by Qatar University Graduate Assistant. The funder played no role in study design, data collection, analysis and interpretation of data, or the writing of this manuscript. 

\section*{Competing interests}
All authors declare no financial or non-financial competing interests. 

\section*{Author contributions}
JK performed data cleaning, implemented algorithms for data analysis, and evaluated the proposed algorithm's performance. YA assisted in implementing the algorithm. JK was the major contributor to the manuscript writing process. MS supervised the project, assisted in implementing the algorithm, evaluated its performance, and reviewed and edited the manuscript. SA acquired funding for the project, supervised it, and provided assistance in reviewing and editing. 

\section*{Data Availability}
The dataset will be made available upon request.

 \bibliographystyle{elsarticle-num} 
\bibliography{MyCollection}% common bib file


% ... (content of your appendix with hyperparameters)



% \end{thebibliography}
\end{document}